\documentclass[10pt,conference]{IEEEtran}  % Comment this line out if you need a4paper

\IEEEoverridecommandlockouts                              % This command is only needed if 
\usepackage{graphics} % for pdf, bitmapped graphics files
\usepackage{epsfig} % for postscript graphics files
\usepackage{times} % assumes new font selection scheme installed
\usepackage{amsmath} % assumes amsmath package installed
\usepackage{amssymb}  % assumes amsmath package installed
\usepackage{color}
\usepackage{algorithm}%
\usepackage{algorithmicx}%
\usepackage{algpseudocode}%
\usepackage{threeparttable}
\usepackage{multirow}
\usepackage{graphicx}
\usepackage{subfigure}

\usepackage{cite}
\usepackage[dvipsnames]{xcolor}
\usepackage{adjustbox}
\usepackage{array}
\usepackage{makecell}
\usepackage{adjustbox}
\usepackage{url}

\definecolor{paleorange}{rgb}{0.85, 0.45, 0.1} % darker orange-brown
\definecolor{skyblue}{rgb}{0.1, 0.45, 0.8}     % darker blue
\definecolor{palegreen}{rgb}{0.1, 0.6, 0.3}    % darker green

\title{\LARGE \bf
}

\title{xFODE+: Explainable Type-2 Fuzzy Additive ODEs for Uncertainty Quantification \\
\thanks{This work was supported by MathWorks\textsuperscript{\textregistered} in part by a Research Grant.
awarded to T. Kumbasar. Any opinions, findings, conclusions, or recommendations expressed in this paper are those of the authors and do not necessarily reflect the views of MathWorks, Inc.}
\author{Ertuğrul Keçeci and Tufan Kumbasar}% <-this % stops a space
\thanks{Ertuğrul Keçeci and Tufan Kumbasar are with AI and Intelligent Systems Laboratory, Istanbul Technical University, 34469, Istanbul, Türkiye  {\tt\small kececie@itu.edu.tr, kumbasart@itu.edu.tr}}%
}

\begin{document}

\maketitle
\thispagestyle{empty}
\pagestyle{empty}

%%%%%%%%%%%%%%%%%%%%%%%%%%%%%%%%%%%%%%%%%%%%%%%%%%%%%%%%%%%%%%%%%%%%%%%%%%%%%%%%
\begin{abstract}
Recent advances in Deep Learning (DL) have boosted data-driven System Identification (SysID), but reliable use requires Uncertainty Quantification (UQ) alongside accurate predictions. Although UQ-capable models such as Fuzzy ODE (FODE) can produce Prediction Intervals (PIs), they offer limited interpretability. We introduce Explainable Type-2 Fuzzy Additive ODEs for UQ (xFODE+), an interpretable SysID model which produces PIs alongside point predictions while retaining physically meaningful incremental states. xFODE+ implements each fuzzy additive model with Interval Type-2 Fuzzy Logic Systems (IT2-FLSs) and constraints membership functions to the activation of two neighboring rules, limiting overlap and keeping inference locally transparent. The type-reduced sets produced by the IT2-FLSs are aggregated to construct the state update together with the PIs. The model is trained in a DL framework via a composite loss that jointly optimizes prediction accuracy and PI quality. Results on benchmark SysID datasets show that xFODE+ matches FODE in PI quality and achieves comparable accuracy, while providing interpretability.
\end{abstract}

%%%%%%%%%%%%%%%%%%%%%%%%%%%%%%%%%%%%%%%%%%%%%%%%%%%%%%%%%%%%%%%%%%%%%%%%%%%%%%%%
\section{Introduction}

System Identification (SysID) has increasingly adopted Deep Learning (DL) to better represent nonlinearities \cite{pillonetto2025deep, dai2024deep, rnn_sysid}. A notable approach is Neural Ordinary Differential Equations (NODEs), where Neural Networks (NNs) define state derivatives \cite{rahman2022neural,gashi2025system}. Several extensions have enhanced the efficiency and interpretability of the NODE \cite{node_sysid, gashi2025system, bottcher2023gradient}. 

Fuzzy Ordinary Differential Equations (FODEs) replace NNs in NODEs with Fuzzy Logic Systems (FLSs) to improve interpretability \cite{guven2025fuzzy}. Yet, their rules are learned over high-dimensional antecedent spaces, and the lack of explicit Partitioning Strategies (PSs) leads to highly overlapping Gaussian Fuzzy Sets (FSs). As a result, the interpretability of FLSs is largely undermined, and FODEs effectively behave as black-box models. Beyond rule-level interpretability, both NODEs and FODEs face additional challenges arising from state representation. Because states are often unmeasured and must be estimated \cite{pillonetto2025deep, dai2024deep}, the learned representations may lack clear physical meaning. Furthermore, both frameworks offer little insight into the role of each state dimension in the dynamics. Additive modeling structures can improve interpretability by isolating each dimension’s effect without giving up flexibility \cite{mariotti2023exploring,yang2021gami, gokmen2025fame}. Explainable FODE (xFODE) \cite{Kececi2026xFODE} addresses interpretability challenges of NODE/FODE by using physically meaningful states, an additive modeling for state derivatives, and PSs in the antecedent space. However, interpretability alone is insufficient for reliability, motivating the incorporation of Uncertainty Quantification (UQ).

\begin{figure}[t] 
\includegraphics[width=\linewidth]{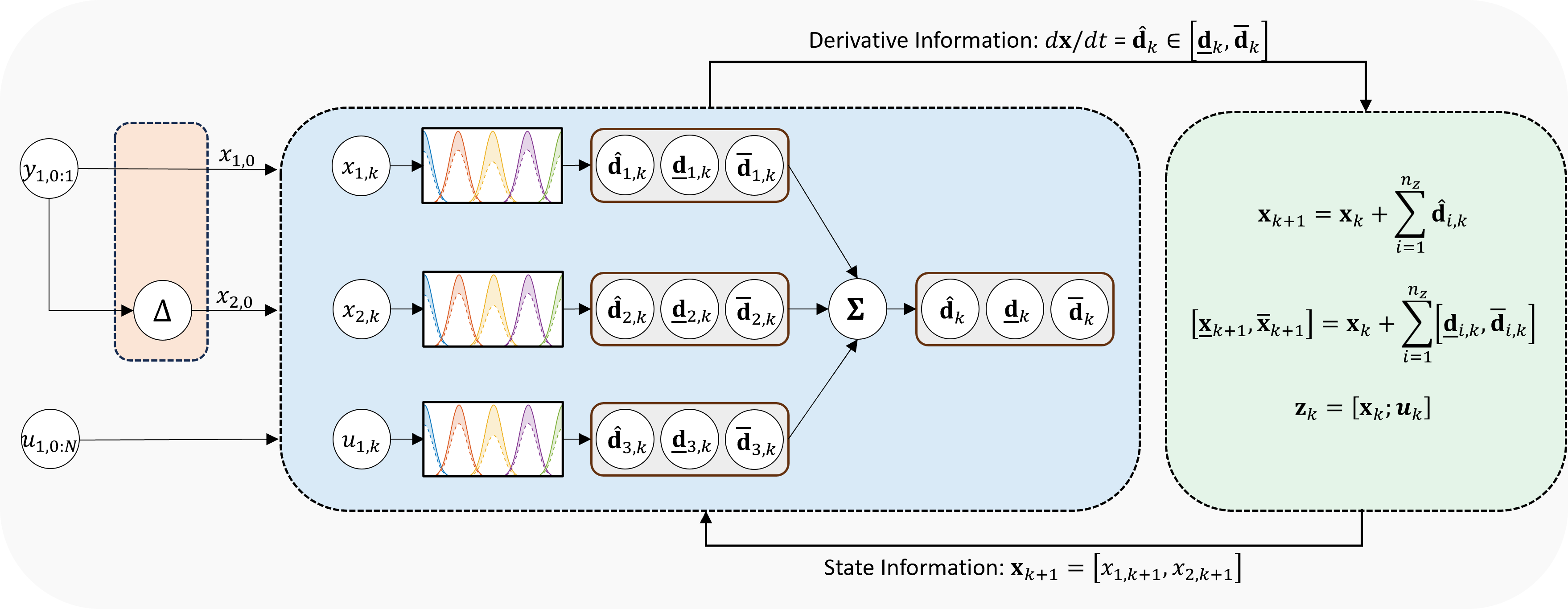}
\caption{Illustration of xFODE+ in a SISO setup with two states: Initial states are obtained from the \textcolor{paleorange}{State Representation} block, defined as $\mathbf{x}_k = [y_k, \Delta y_k]$. Then, together with the input $u(k)$, they are fed to the \textcolor{skyblue}{IT2-FLSs} to generate the derivative information, which is processed by the \textcolor{palegreen}{ODE} block to compute the next states and the PI, i.e. $\mathbf{x}_{k+1} \in \left[\underline{\mathbf{x}}_{k+1},\overline{\mathbf{x}}_{k+1}\right]$.}
    \label{fig:xfode+}
\end{figure}

For reliable deployment of SysID models, UQ augments predictions with confidence information, commonly via Prediction Intervals (PIs) that bound unseen targets with confidence level $\delta$ \cite{vovk2005algorithmic,saleh,uq_survey}. FODE~\cite{guven2025fuzzy} incorporates UQ by modeling the antecedent space with Interval Type-2 (IT2) FSs, where each Membership Function (MF) is defined by a Footprint of Uncertainty (FOU) bounded by Lower and Upper MFs (LMF/UMF). Yet, in high-dimensional spaces, significant overlap among IT2-FSs can limit the interpretability of UQ.

This paper proposes \emph{Explainable FODE+} (xFODE+), an interpretable data-driven SysID model that augments xFODE with UQ by producing PIs, as illustrated in Fig.~\ref{fig:xfode+}. In xFODE+, states are represented in an incremental form to preserve the physically grounded temporal structure. To provide UQ, xFODE+ replaces Type-1 (T1) FLSs used in xFODE with IT2-FLSs. By aggregating Type-Reduced Sets (TRS) generated by each IT2-FLS, xFODE+ produces a state update together with its PI. For interpretability, we present PSs to the IT2-FLSs by constraining the MFs such that only two neighboring rules can fire at a time, limiting MF overlap and keeping inference locally transparent. We train xFODE+ end-to-end in a DL framework by minimizing a composite loss that combines predictive accuracy with PI quality. Comparisons with NODE, FODE, and xFODE on benchmark datasets show that xFODE+ achieves FODE-level PI quality while preserving NODE-level accuracy alongside interpretability.

\section{Neural and Fuzzy ODE Networks}

Consider a non-autonomous partially observable nonlinear system described by
\begin{equation}
\label{dynamics}
\begin{aligned}
    \dot{\mathbf{x}}(t) &= d(\mathbf{x}(t),\mathbf{u}(t)), \ \mathbf{x}(t_0) = \mathbf{x}_0, \
    \mathbf{y}(t) = h(\mathbf{x}(t))
\end{aligned}
\end{equation}
where $\mathbf{x}(t) \in \mathbb{R}^{n_x}$ denotes the state vector, $\mathbf{u}(t) \in \mathbb{R}^{n_u}$ the input vector, and $\mathbf{y}(t) \in \mathbb{R}^{n_y}$ the outputs. $d(\cdot)$ defines the nonlinear dynamics, whereas $h(\cdot)$ maps states to outputs.

To model dynamics in \eqref{dynamics}, the evolution of states of NODEs and FODEs is governed by a parameterized vector field $d^{\mathrm{M}}(\cdot)$, where $\mathrm{M} \in \{\mathrm{NN},\,\mathrm{T1},\,\mathrm{IT2}\}$, defined by
\begin{equation} \label{neural_dynamics}
    \dot{\mathbf{x}}(t) = d^\mathrm{M}\left(\mathbf{x}(t), \mathbf{u}(t); {\boldsymbol{\theta}} \right)
\end{equation}
where $\boldsymbol{\theta}$ denotes the Learnable Parameters (LPs). Given initial condition $\mathbf{x}(t_0)$ the state at $t_1$ is defined as
\begin{equation}
\label{xtraj}
    \mathbf{x}(t_1) = \mathbf{x}(t_0) + \int_{t_0}^{t_1} d^\mathrm{M}(\mathbf{x}(\tau),\, \mathbf{u}(\tau); \boldsymbol{\theta})\, d\tau.
\end{equation}
Using Euler integration yields the discrete-time form of \eqref{dynamics}:
\begin{equation}
\label{discnonlin}
\begin{aligned}
    \mathbf{x}_{k+1} = \mathbf{x}_k + d^\mathrm{M}(\mathbf{z}_k; \boldsymbol{\theta}), \
    \mathbf{y}_k = h(\mathbf{x}_k),
\end{aligned}
\end{equation}
where $\mathbf{z}_k=[\mathbf{x}_k;\mathbf{u}_k]\in\mathbb{R}^{n_z}$ with $n_z=n_x+n_u$. In particular, for xFODE \cite{Kececi2026xFODE}, the state update in \eqref{discnonlin} is in additive form:
\begin{equation}
    \mathbf{x}_{k+1} = \mathbf{x}_k + \sum_{i=1}^{n_z} d^\mathrm{T1}_i(\mathbf{z}_{i,k}; \boldsymbol{\theta}_i)
\end{equation}
When FODE is defined with IT2-FLSs, it outputs the TRS $\left[\underline{\mathbf{d}}_{k},\,\overline{\mathbf{d}}_{k}\right]$ alongside crisp output \cite{guven2025fuzzy}:
\begin{equation}
\label{it2-output}
\left[\hat{\mathbf{d}}_{k},\,\underline{\mathbf{d}}_{k},\,\overline{\mathbf{d}}_{k}\right] 
= d^{\mathrm{IT2}}(\mathbf{z}_{k};\boldsymbol{\theta}),
\end{equation}
Accordingly, IT2-FODE yields a PI for the next state:
\begin{equation}
\label{it2_fode_band}
\big[\underline{\mathbf{x}}_{k+1},\,\overline{\mathbf{x}}_{k+1}\big]
=
\mathbf{x}_{k}+\Big[\underline{\mathbf{d}}_{k},\ \overline{\mathbf{d}}_{k}\Big],
\end{equation}
while the state update in \eqref{discnonlin} uses $\hat{\mathbf{d}}_{k} = \left(\underline{\mathbf{d}}_{k} + \overline{\mathbf{d}}_{k}\right)/\,2$. We refer the reader to \cite{pillonetto2025deep, guven2025fuzzy,Kececi2026xFODE} for more details.

\section{xFODE+ Framework}

This section introduces xFODE+, which augments xFODE with UQ capability via PIs (Fig.~\ref{fig:xfode+}).

\subsection{Interpretable State Representation}\label{staterep}

A standard approach for constructing the state vector $\mathbf{x}_k$ of NODE/FODE is directly from measured outputs \cite{guven2025fuzzy}:
\begin{flalign} \label{sr0}
  \text{SR0: } && \mathbf{x}_k = \mathbf{y}_k, &&
\end{flalign}
which is an interpretable state, as $\mathbf{x}_k$ corresponds to physical output variables. Yet, it does not encode temporal dependencies, which are essential for capturing complex dynamics.

To encode temporal dependencies while preserving interpretability, we suggest defining $\mathbf{x}_k$ in the \emph{incremental form}:
\begin{flalign}
\label{incremental}
   \text{SR1: }&& \mathbf{x}_k = [\,\mathbf{y_k},\,\Delta \mathbf{y}_k,\, \dots,\, \Delta^m \mathbf{y}_k\,]^T&&
\end{flalign}
with $m$ denoting the number of difference orders and $\Delta^j \mathbf{y}_k = \Delta^{j-1}\mathbf{y}_k - \Delta^{j-1}\mathbf{y}_{k-1}, j = 1, \dots, m$. This representation emphasizes temporal changes, with each term corresponding to a physically meaningful quantity, such as velocity or acceleration of the outputs.

% \begin{equation}
%     \Delta^j \mathbf{y}_k = \Delta^{j-1}\mathbf{y}_k - \Delta^{j-1}\mathbf{y}_{k-1}, \quad j = 1, \dots, m.
% \end{equation}

\subsection{Additive ODE Dynamics}

The state derivative in xFODE+ is represented additively.
\begin{equation}
\label{xfode}
    \mathbf{x}_{k+1} = \mathbf{x}_k + \sum_{i=1}^{n_z} \hat{\mathbf{d}}_{i,k},
\end{equation}
where $\hat{\mathbf{d}}_{i,k} \in \mathbb{R}^{n_x}$ represents the contribution of $z_{i,k}$. Each contribution is produced by $f_i(\cdot)$, which outputs both a point prediction and a corresponding TRS:
\begin{equation}
\label{fi_output}
\left[\hat{\mathbf{d}}_{i,k},\,\underline{\mathbf{d}}_{i,k},\,\overline{\mathbf{d}}_{i,k}\right] = f_i(z_{i,k},\boldsymbol{\theta}_i).
\end{equation}
To obtain a PI for the next state, we define:
\begin{equation}
\label{it2_band_add}
\big[\underline{\mathbf{x}}_{k+1},\,\overline{\mathbf{x}}_{k+1}\big]
=
{\mathbf{x}}_{k}+\Big[\sum_{i=1}^{n_z}\underline{\mathbf{d}}_{i,k},\ \sum_{i=1}^{n_z}\overline{\mathbf{d}}_{i,k}\Big].
\end{equation}
This additive structure allows xFODE+ to provide input-wise interpretability while generating PIs.

\begin{figure*}[t]
        \centering
        \subfigure[Learned GaussMF if there is no PS employed]
        {
        \includegraphics[width=0.35\textwidth,height=0.21\textheight,keepaspectratio]{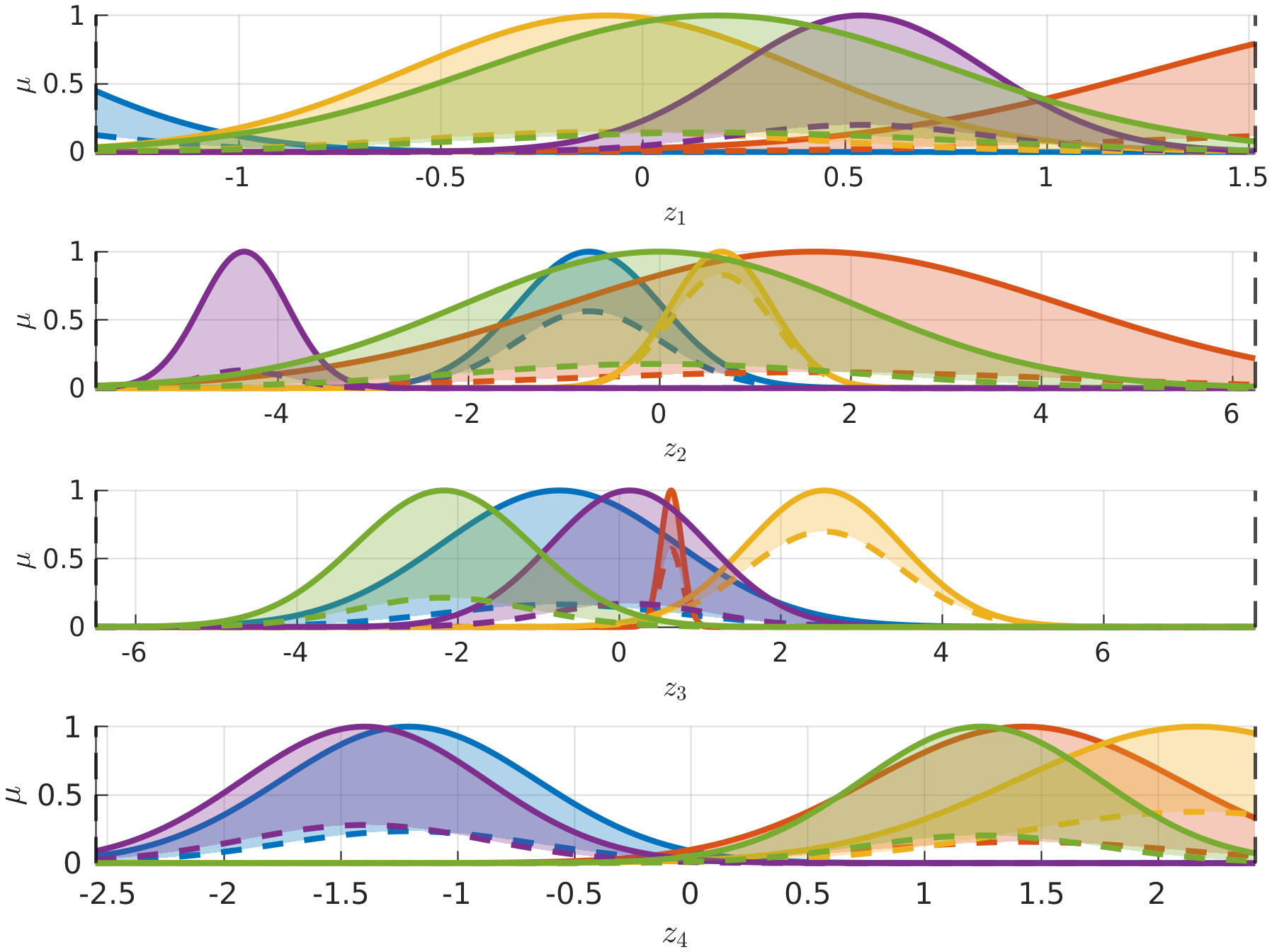}\label{fig:gaussmf}

        }
        \subfigure[Learned TriMFs via PS1]
        {
        \includegraphics[width=0.35\textwidth,height=0.21\textheight,keepaspectratio]{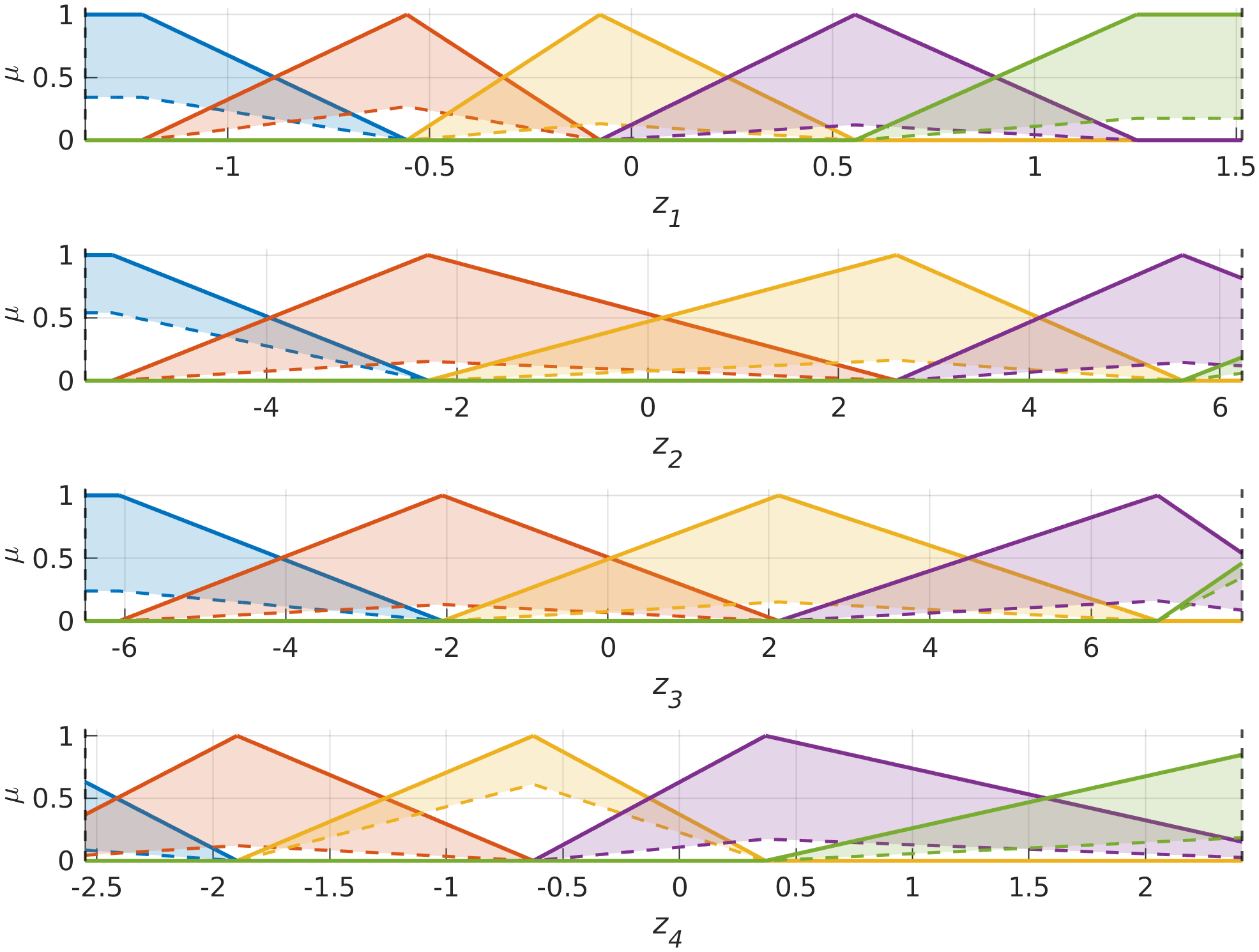}\label{fig:trimf}
        
        }

        \subfigure[Learned Gauss2MF via PS2]
        {
        \includegraphics[width=0.35\textwidth,height=0.21\textheight,keepaspectratio]{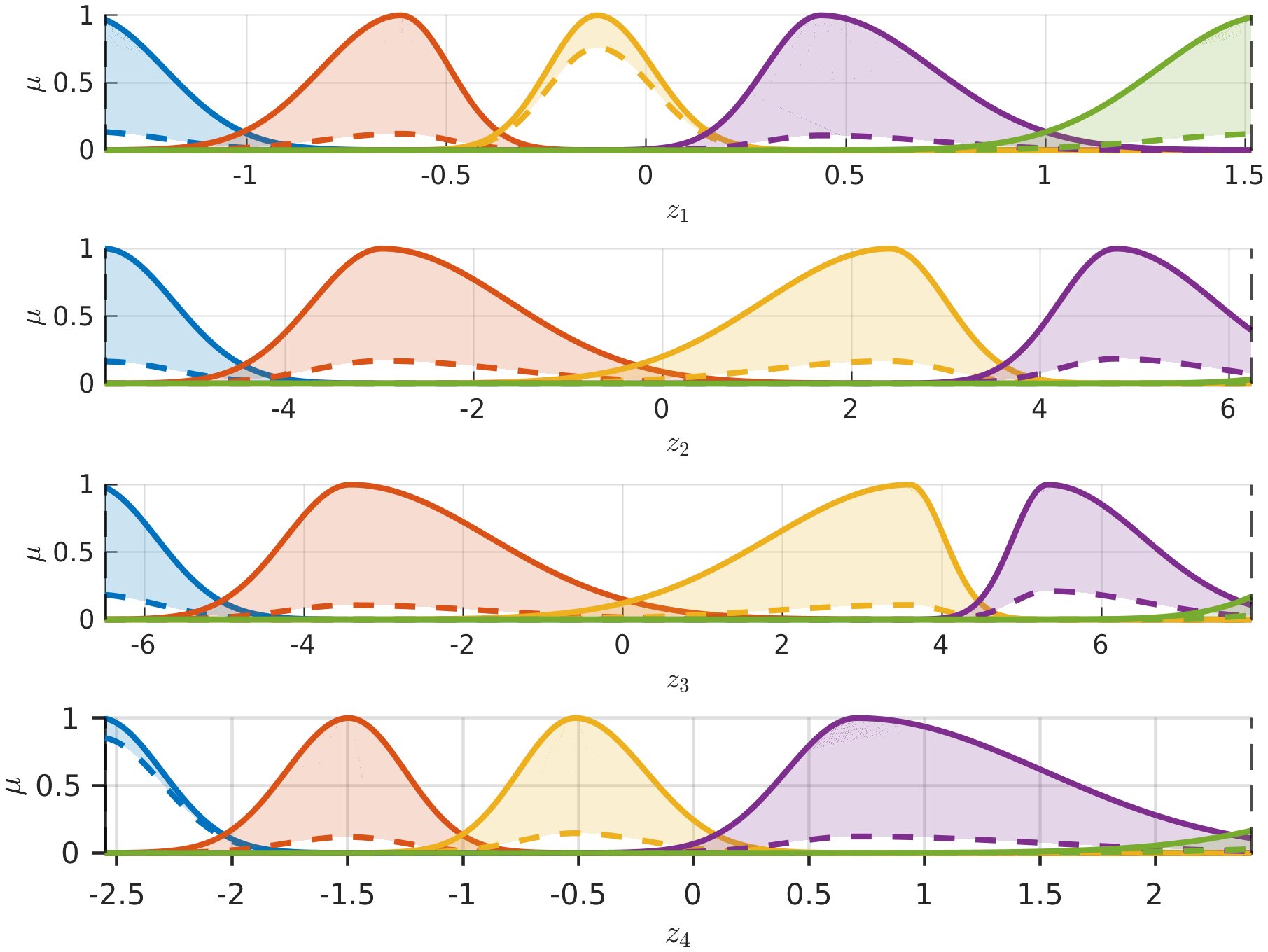}\label{fig:gauss2mf-1}
        }
        \subfigure[Learned Gauss2MF via PS3]
        {
        \includegraphics[width=0.35\textwidth,height=0.21\textheight,keepaspectratio]{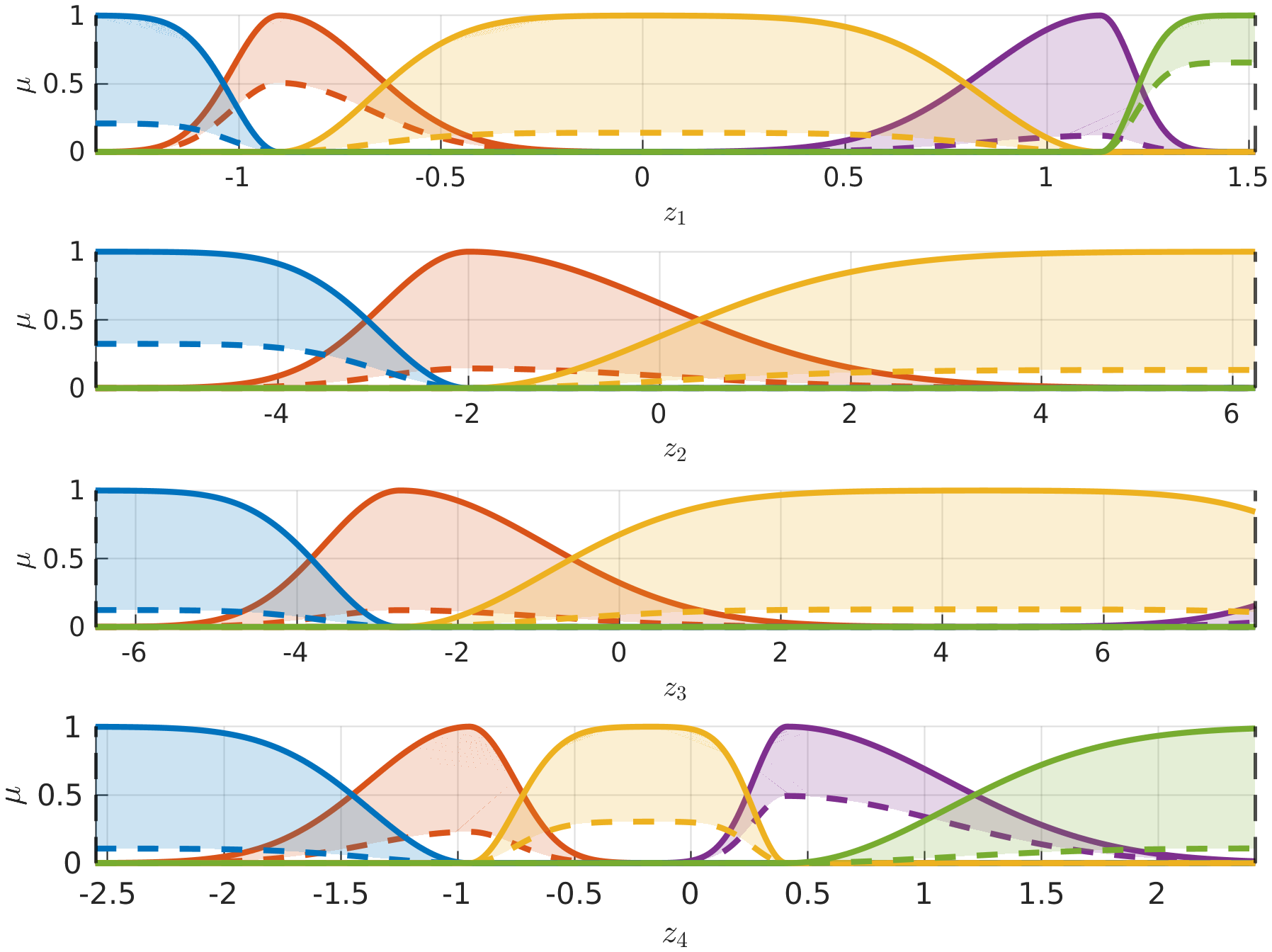}\label{fig:gauss2mf-2}
        
        }
        \caption{Visualization of learned IT2-FSs on the MR Damper dataset (single seed). States are defined with the incremental form (SR1), thus $\mathbf{z} = [\mathbf{y}; \Delta\mathbf{y}; \Delta^2\mathbf{y};\mathbf{u}]$. Each combined input dimension $z_i$ is interpretable: \(z_1\) output position, \(z_2\) output velocity, \(z_3\) output acceleration, and \(z_4\) control input.}
        \label{fig:MFs}
\end{figure*}

\subsection{Interpretable Mapping Design} \label{antdesign}

In xFODE+, each $f_i(.)$ in \eqref{fi_output} is implemented as a single-input IT2-FLS that maps $f_i: \mathbb{R} \to \mathbb{R}^{n_x} \times \mathbb{R}^{n_x} \times \mathbb{R}^{n_x}$. For notational simplicity, we drop the indices $i$ and $k$.

The IT2-FLS has $P$ rules ($p = 1, \dots, P$), defined as
\begin{equation}
R_p:\ \text{If $z$ is $\hat{A}_p$ then $\hat{\mathbf d}$ is $\mathbf d_p(z)$},
\end{equation}
where $\mathbf{d}_p(z)=[d_{p,1}(z),\dots,d_{p,n_x}(z)]^T$ and
$d_{p,o}(z)=a_p^o z + a_{p,0}^o$, $o=1,\dots,n_x$. The crisp output is calculated as $\hat{\mathbf d} = \big(\underline{{\mathbf d}} + \overline{{\mathbf d}}\big)/2$,  where $\underline{{\mathbf d}}=[\underline{ d}^{(1)},\dots,\underline{ d}^{(n_x)}]^T$ and
$\overline{{\mathbf d}}=[\overline{ d}^{(1)},\dots,\overline{ d}^{(n_x)}]^T$. The TRS is obtained via the KM algorithm, which is performed for each output separately.
\begin{equation}\label{kmtr}
\begin{aligned}
\underline{ d}^{(o)}
&=
\frac{
\sum_{p=1}^{L} \overline{\mu}_p(z)\, d_{p,o}(z)
+
\sum_{p=L+1}^{P} \underline{\mu}_p(z)\, d_{p,o}(z)
}{
\sum_{p=1}^{L} \overline{\mu}_p(z)
+
\sum_{p=L+1}^{P} \underline{\mu}_p(z)
}
\\
\overline{ d}^{(o)}
&=
\frac{
\sum_{p=1}^{R} \underline{\mu}_p(z)\, d_{p,o}(z)
+
\sum_{p=R+1}^{P} \overline{\mu}_p(z)\, d_{p,o}(z)
}{
\sum_{p=1}^{R} \underline{\mu}_p(z)
+
\sum_{p=R+1}^{P} \overline{\mu}_p(z)
}
\end{aligned}
\end{equation}
Here, $L$ and $R$ are the switching points of the KM algorithm. 

In \eqref{kmtr}, $\underline{\mu}_p(z)$ and $\overline{\mu}_p(z)$ are the membership grades of the LMF and UMF. Gaussian MFs (GaussMFs) are commonly used due to their capacity to model nonlinearities.  
\begin{equation} \label{lmf}
\begin{aligned}
    \overline{\mu}_{p}(z) = \text{exp}(-(z-c_p)^2 / 2(\sigma_p)^2), \
    \underline{\mu}_p(z) = h_p\overline{\mu}_p(z)
\end{aligned}
\end{equation}
where $c_p$ and $\sigma_p$ denote the center and standard deviation, respectively, and $h_p \in (0,1]$ defines the height of the LMF and thus the size of the FOU. As shown in Fig. \ref{fig:gaussmf}, if there are no PSs, their overlap can cause multiple rules to be simultaneously fired, reducing interpretability. 

To address this, we propose PSs to sculpt the antecedent space so that only two consecutive rules are activated. As a result, for \(z' \in [c_{p^*},\, c_{p^*+1}]\), the KM is performed with $N^* = 2$, yielding $L = R = 1$ \cite{kumbasar2015robust}, which simplifies \eqref{kmtr} to: 
\begin{equation}\label{kmtrsimp}
\begin{aligned}
\underline{ d}^{(o)} &=
\frac{
\overline{\mu}_{p^*}(z')\, d_{p^*,o}(z')
+
\underline{\mu}_{p^*+1}(z')\, d_{p^*+1,o}(z')
}{
\overline{\mu}_{p^*}(z')
+
\underline{\mu}_{p^*+1}(z')
},
\\
\overline{d}^{(o)} &=
\frac{
\underline{\mu}_{p^*}(z')\, d_{p^*,o}(z')
+
\overline{\mu}_{p^*+1}(z')\, d_{p^*+1,o}(z')
}{
\underline{\mu}_{p^*}(z')
+
\overline{\mu}_{p^*+1}(z')
}.
\end{aligned}
\end{equation}
This ensures a simple mapping while maintaining interpretability. For all presented PSs, we define LMFs as in \eqref{lmf}.

\subsubsection{PS1-Partitioning with Triangular MFs (TriMFs)} TriMFs enable a simple partitioning; the UMFs are set as
\begin{equation}\label{utrimf}
\overline{\mu}_p(z) = \max\Big(0, \min\Big(\frac{z-l_p}{c_p-l_p}, \frac{r_p-z}{r_p-c_p}\Big)\Big)
\end{equation}
where $l_p$, $c_p$, and $r_p$ denote the left feet, center, and right feet. For partitioning, the following couplings are defined:
\begin{align}
\Delta_p^l = c_p - l_p, \quad
\Delta_p^r = r_p - c_p, \nonumber
\\
c_{p+1} = c_p + \Delta_p^r, \quad
\Delta_{p+1}^l = \Delta_p^r.
\end{align}
This coupling ensures that, for any input $z'$, only two consecutive UMFs are activated as shown in Fig. \ref{fig:trimf}.

\subsubsection{PS2-Partitioning with Two-Sided Gaussian MFs (Gauss2MFs)}
To provide flexibility and smoothness, we employ two-sided Gaussian MFs (Gauss2MFs) as UMFs:
\begin{equation}
\label{gauss2mf}
\overline{\mu}_{p}(z)=\left\{\begin{array}{l}
\exp \left(-\left(z-c_p\right)^2 / 2\left(\sigma_p^l\right)^2\right), \text { if } z \leq c_p \\
\exp \left(-\left(z-c_p\right)^2 / 2\left(\sigma_p^r\right)^2\right), \text { if } z>c_p
\end{array}\right.
\end{equation}
where $\sigma^l_p$ and $\sigma^r_p$ are the left and right standard deviations, respectively. For interpretability, we parameterize it as: 
\begin{equation}
    \label{centers}
        c_{p+1} = c_p + 4\sigma^r_p,\quad
        \sigma^r_p = \sigma^l_{p+1}       
\end{equation}
This results in a partition as shown in Fig. \ref{fig:gauss2mf-1}. Note that we assume $\overline{\mu}_{p}(z') \approx 0$ when $|z'|>4\sigma^r_p$ \cite{gokmen2025fame}.  

\subsubsection{PS3-Partitioning with Complementary Gauss2MFs} 
In PS2, while Gauss2MFs offer asymmetric flexibility, the $4\sigma$ center spacing in \eqref{centers} may result in weak overlap between adjacent UMFs. Thus, to increase the overlap, we construct odd-indexed UMFs as local complements of neighboring even-indexed UMFs. Let even indexed \(A_{2q}\) (\(q=1,2,\dots, \left\lceil (P-3)/2 \right\rceil\)) have UMFs $\overline{\mu}_{2q}(z)$ defined by \eqref{gauss2mf}. The midpoint between consecutive even-indexed centers is:
\begin{equation}
{c}_{2q+1}^{\text{mid}} = ({c_{2q} + c_{2(q+1)}})/{2}.
\end{equation}
The odd indexed UMFs $\overline{\mu}_{2q+1}(z)$ are then defined as
\begin{equation}
\overline{\mu}_{2q+1}(z) =
\begin{cases}
0, & z < c_{2q},\\
1-\overline{\mu}_{2q}(z), & c_{2q} \le z \le {c}_{2q+1}^{\text{mid}},\\
1-\overline{\mu}_{2(q+1)}(z), & {c}_{2q+1}^{\text{mid}} \le z \le c_{2(q+1)},\\
0, & z > c_{2(q+1)},
\end{cases}
\end{equation}
with \(\overline{\mu}_1(z) = 1-\overline{\mu}_2(z), z < c_2\) and when $P$ is odd, \(\overline{\mu}_P(z) = 1-\overline{\mu}_{P-1}(z), z > c_{P-1}\). This UMF design guarantees that only two UMFs are active as shown in Fig. \ref{fig:gauss2mf-2}. 

% \subsection{Inference of xFODE+}

% To simulate the system dynamics over a prediction horizon $N$, xFODE+ defines $\hat{\mathbf z}_k$ by using its predicted state $\hat{\mathbf x}_k$ at time step $k$. Thus, each $f_i(\cdot)$ produces:
% \begin{equation}
% \label{it2fls}
% \big[\hat{\mathbf d}_{i,k},\,\underline{\mathbf d}_{i,k},\,\overline{\mathbf d}_{i,k}\big]
% = f_i(z_{i,k},\boldsymbol{\theta}_i),
% \end{equation}
% where the TR bounds $\big[\underline{\mathbf d}_{i,k},\,\overline{\mathbf d}_{i,k}\big]$ are computed via \eqref{kmtrsimp} while the corresponding crisp output is obtained via \eqref{point}. Then, the derivative-like updates and their TR bounds are aggregated as:
% \begin{equation}
% \label{agg}
%     \left[\hat{{\mathbf d}}_{k},\,\underline{{\mathbf d}}_{k},\,\overline{\mathbf d}_{k}\right] = \sum_{i=1}^{n_z}\left[\hat{{\mathbf d}}_{i,k},\,\underline{{\mathbf d}}_{i,k},\,\overline{\mathbf d}_{i,k}\right]
% \end{equation}
% The state prediction $\hat{\mathbf x}_{k+1}$ and its PI are propagated as:
% \begin{equation}
% \label{predandpi}
% \big[\hat{\mathbf x}_{k+1},\,\underline{{\mathbf x}}_{k+1},\,\overline{{\mathbf x}}_{k+1}\big]
% =
% \hat{\mathbf x}_{k} + \big[\hat{\mathbf d}_{k},\,\underline{{\mathbf d}}_{k},\,\overline{\mathbf d}_{k}\big].
% \end{equation}
% xFODE+ yields predicted state trajectories $\{\hat{\mathbf x}_k\}_{k=1}^N$ and the corresponding PIs $\{C_k\}_{k=1}^N$ where $C_k = \left[\underline{{\mathbf x}}_k,\overline{{\mathbf x}}_k]\right]$.

\section{DL Framework for xFODE+} \label{dlfram}

We next introduce a DL framework to train xFODE+. Algorithm \ref{alg:alg1}\footnote[1]{MATLAB implementation. [Online]. Available: \newline \url{https://github.com/ertugrulkececi/xfodeplus}} outlines the training procedure and Algorithm \ref{alg:alg2} describes the simulation inference. We consider a dataset $D = \{\mathbf{u}_k, \mathbf{y}_k\}_{k=1}^K$. After defining $\mathbf{x}_k$, the resulting input–state trajectories for $j \in[1, \ldots, B$ are:
\begin{equation}
S=\left[\left(\mathbf{u}_0^{\{j\}}, \mathbf{x}_0^{\{j\}}\right), \ldots,\left(\mathbf{u}_N^{\{j\}}, \mathbf{x}_N^{\{j\}}\right)\right]
\end{equation}
Here, $N$ is the roll-out while $B$ is the number of trajectories.

\subsection{Loss Function}
xFODE+ learns the underlying system dynamics by minimizing a composite loss function over $S$:
\begin{equation}
\label{comploss}
    \min_{\boldsymbol{\theta \in \mathcal{C}}} L_C = L_A + L_{UQ}
\end{equation}
where the accuracy loss 
\begin{equation}
\label{l1loss}
    L_A = \frac{1}{B}\sum_{j=1}^B \sum_{k=1}^N |\mathbf{x}_k^{\{j\}} - \hat{\mathbf{x}}_k^{\{j\}}|
\end{equation}
% with $\hat{\mathbf{x}}_k^{\{j\}}$ representing the predicted states by xFODE+:
% \begin{equation}
%     \hat{\mathbf{x}}_{k} = \hat{\mathbf{x}}_{k-1} + \sum_{i=1}^{n_z} \hat{\mathbf d}_{i,k-1}.
% \end{equation}

\begin{algorithm}[t]
\caption{Training steps of xFODE+} \label{alg:alg1}
\begin{algorithmic}[1]
\State \textbf{Input:} Dataset $D = \{\mathbf{u}_k, \mathbf{y}_k\}_{k=1}^K$, prediction horizon $N$, number of rules $P$, mini-batch size $mbs$, number of epochs $E$, number of difference orders $m$, desired quantile levels $\{\underline{\tau},\overline{\tau}\}$
\State \textbf{Output:} LP set ${\theta}$
\State Initialize ${\theta}$;
\State Construct $\mathbf{x}_k$ via \eqref{incremental};
\State Build $S = (\mathbf{u}^{\{j\}}_{1:N},\mathbf{x}^{\{j\}}_{1:N})^{B}_{j=1}$;
\For{$e = 1 \text{ to } E$}
\For{\textbf{each } $mbs$ in $B$} 
    \State $\text{Select a mini-batch } [ \mathbf{u}^{\{j\}}_{0:N},\mathbf{x}^{\{j\}}_{0:N}]^{mbs}_{j=1}$
    \State $\big[\hat{\mathbf x}^{\{j\}}_{1:N},\,\underline{{\mathbf x}}^{\{j\}}_{1:N},\,\overline{{\mathbf x}}^{\{j\}}_{1:N}\big] \leftarrow \text{xFODE+}([\mathbf{x}^{\{j\}}_{0}, \mathbf{u}^{\{j\}}_{0:N}]; {{\theta}})$
    \State Compute $L_C$ and the gradient ${\partial L_C}/{\partial {\theta}} $ 
    \State Update ${\theta}$ via a DL optimizer, e.g., Adam 
\EndFor
\EndFor
\State ${\theta}^* = {\arg \min }(L_C)$
\State \textbf{Return} $\theta = {\theta}^{*}$
\end{algorithmic}
\end{algorithm}

\begin{algorithm}[t]
\caption{Inference of xFODE+ for a trajectory} \label{alg:alg2}
\begin{algorithmic}[1]
\State \textbf{Input:} Initial state $\mathbf{x_0}$, input sequence $\mathbf{u}_{0:N}$, LP set ${\theta}$, prediction horizon $N$
\State \textbf{Output:} Predicted state trajectory $\hat{\mathbf{x}}_{1:N}$
\State Initialize $\big[\hat{\mathbf x},\,\underline{{\mathbf x}},\,\overline{{\mathbf x}}\big] \gets \varnothing$, $\hat{\mathbf{x}}_0 = \mathbf{x}_0$
\For{$k = 0$ \textbf{to} $N-1$}
    \State $\mathbf{z_k} \leftarrow [\hat{\mathbf{x}}_{k}, \mathbf{u}_{k}]$
    \State $\big[\hat{\mathbf d}_{k},\,\underline{\mathbf d}_{k},\,\overline{\mathbf d}_{k}\big] \gets \sum_{i=1}^{n_z} f_i(z_{i,k},\boldsymbol{\theta}_i)$
    \State $\big[\hat{\mathbf{x}}_{k+1},\, \underline{\mathbf{x}}_{k+1},\,\overline{\mathbf{x}}_{k+1}\big] \gets \hat{\mathbf{x}}_{k} +\big[\hat{\mathbf d}_{k},\,\underline{\mathbf d}_{k},\,\overline{\mathbf d}_{k}\big]$
   \State $\big[\hat{\mathbf x},\,\underline{\mathbf x},\,\overline{\mathbf x}\big] \gets \big[\hat{\mathbf x},\,\underline{\mathbf x},\,\overline{\mathbf x}\big]\cup \big[\hat{\mathbf x}_{k+1},\,\underline{\mathbf x}_{k+1},\,\overline{\mathbf x}_{k+1}\big]$
    \State $\hat{\mathbf{x}}_{k} \gets \hat{\mathbf{x}}_{k+1}$
\EndFor
\State \textbf{Return} $\big[\hat{\mathbf x},\,\underline{{\mathbf x}},\,\overline{{\mathbf x}}\big]$
\end{algorithmic}
\end{algorithm}

To train xFODE+ to produce valid PIs at each time step, the TRS $C_k = \left[\underline{\mathbf{x}}_k,\overline{\mathbf{x}}_k\right]$. are required to satisfy the marginal coverage condition $\text{I\kern-0.20em P}\{{\mathbf x}_k \in C_k\}\ge \delta$. To learn such PIs, we employ the tilted (pinball) loss. For a residual $e$ and quantile level $\tau \in (0,1)$, the tilted loss is defined as $\rho_{\tau}(e) = \max(\tau e, (\tau - 1)e)$. The resulting UQ loss is defined as
\begin{equation}
\label{uqloss}
L_{UQ}=\frac{1}{B}\sum_{j=1}^{B}\sum_{k=1}^{N}
\Big(\rho_{\underline{\tau}}(\underline{e}^{(j)}_{k})+\rho_{\overline{\tau}}(\overline{e}^{(j)}_{k})\Big),
\end{equation}
with  $\underline{e}_k^{\{j\}} = \mathbf{x}_k^{\{j\}} - \underline{\mathbf{x}}_k^{\{j\}}$ and $\overline{e}_k^{\{j\}} = \mathbf{x}_k^{\{j\}} - \overline{\mathbf{x}}_k^{\{j\}}$. Here, $\underline{\tau}$ and $\overline{\tau}$ denote the lower and upper quantile levels. For a given $\delta$, we set $\underline{\tau} = (1-\delta)/2$ and $\overline{\tau} = 1 - (1-\delta)/2$. 
% The residuals in~\eqref{uqloss} are computed as
% \begin{equation}
% \begin{aligned}
%     \underline{e}_k^{\{j\}} = \mathbf{x}_k^{\{j\}} - \underline{\mathbf{x}}_k^{\{j\}}, \quad \overline{e}_k^{\{j\}} = \mathbf{x}_k^{\{j\}} - \overline{\mathbf{x}}_k^{\{j\}}
% % \\
% %     \left[\underline{\mathbf{x}}_k^{\{j\}},\,\overline{\mathbf{x}}_k^{\{j\}}\right] = \hat{\mathbf{x}}_{k-1}^{\{j\}} + \sum_{i=1}^{n_z} \left[\underline{\mathbf d}_{i,k-1},\,  \overline{\mathbf d}_{i,k-1}\right].
% \end{aligned}
% \end{equation}
\subsection{Parameterization Tricks}
The complete set of LPs is defined as $\boldsymbol{\theta} = \{\boldsymbol{\theta}_i\}^{n_z}$ where each $\boldsymbol{\theta}_i$ is parameterized according PS:
\begin{itemize}
    \item For PS1: $\boldsymbol{\theta}_i = \{c_1,\Delta_1^l,\boldsymbol{\Delta}^r, \mathbf{h},\mathbf{d}\}$ where $c_1$ and $\Delta_1^l$ are scalar, $\boldsymbol{\Delta}^r \in \mathbb{R}^{P \times 1}$, $\mathbf{h} \in \mathbb{R}^{P \times 1}$, and $\mathbf{d} \in \mathbb{R}^{2Pn_x \times 1}$, 
    \item For PS2/ PS3: $\boldsymbol{\theta}_i = \{c_1,\sigma_1^l,\boldsymbol{\sigma}^r, \mathbf{h},\mathbf{d}\}$ where $c_1$ and $\sigma_1^l$ $\sigma$ are scalar, $\boldsymbol{\sigma}^r \in \mathbb{R}^{P \times 1}$, $\mathbf{h} \in \mathbb{R}^{P \times 1}$, and $\mathbf{d} \in \mathbb{R}^{2Pn_x \times1}$
\end{itemize}
During training, it is necessary to enforce $\sigma_p > 0$ or $\Delta_p > 0$, $\forall p$ \cite{beke2022more}. This is achieved via softplus reparameterization, i.e., $\Delta_p=\log(1+e^{\Delta_p'})$ for TriMFs and $\sigma_p=\log(1+e^{\sigma_p'})$ for Gauss2MFs, while $h_p$ is constrained to $h_p \in (0.1,1)$ using a scaled sigmoid mapping $h_p = 0.1 + 0.9(1+e^{-h_p'})^{-1}$ to limit the FOU size and improve computational stability.

\section{Performance Analysis}
We evaluate the SysID performance of xFODE+ on three benchmark datasets: Hair Dryer, MR Damper, and Steam Engine. All datasets are normalized prior to training and split into training/test as in \cite{Kececi2026xFODE}. Experiments are conducted in MATLAB and repeated over 20 independent runs. All models are evaluated in simulation (multi-step prediction) mode with a roll-out length of $N=20$. For UQ, a target coverage of $99\%$ is defined. Model accuracy is quantified using RMSE, while the quality of PIs is evaluated using the PI Coverage Percentage (PICP) and the PI Normalized Average Width (PINAW)~\cite{saleh}. 

The following models are considered:
\begin{itemize}
    \item \textbf{NODE:} $d^{\mathrm{NN}}(\cdot)$ is a neural network with two hidden layers of 128 units each and \textit{tanh} activations. The NODE models are trained with $L_A$ as training loss.
    \item \textbf{T1-FODE:} $d^{\mathrm{T1}}(\cdot)$ is a multi-input T1-FLS using GaussMFs. The models are trained with $L_A$ \cite{guven2025fuzzy}.
    \item \textbf{xFODE:} The models are learned with PS1--PS3 and $L_A$ as training loss \cite{Kececi2026xFODE}.
    \item \textbf{IT2-FODE:} $d^{\mathrm{IT2}}(\cdot)$ is a multi-input IT2-FLS, using GaussMFs. The models are trained with $L_C$ \cite{guven2025fuzzy}.
    \item \textbf{xFODE+:} Each IT2-FLS $f_i(\cdot)$ in~\eqref{fi_output} is constructed using PS1--PS3 and trained as described in Section \ref{dlfram}.
    \item \textbf{Additive FODE+:} To assess the effect of PSs, we train xFODE+ using GaussMFs without any PS (AFODE+).
\end{itemize}
We trained all models with both the SR0 and SR1 representations. The hyperparameter of SR1 $m$ is selected via cross-validation on NODE and fixed across all models for fairness. The resulting values are $m=2$ for the Hair Dryer and MR Damper datasets, and $m=1$ for the Steam Engine dataset. All fuzzy models are defined with $P=5$ rules. 

\begin{figure}[t]
        \centering
          \subfigure{   \includegraphics[width=0.46\textwidth,height=0.21\textheight,keepaspectratio]{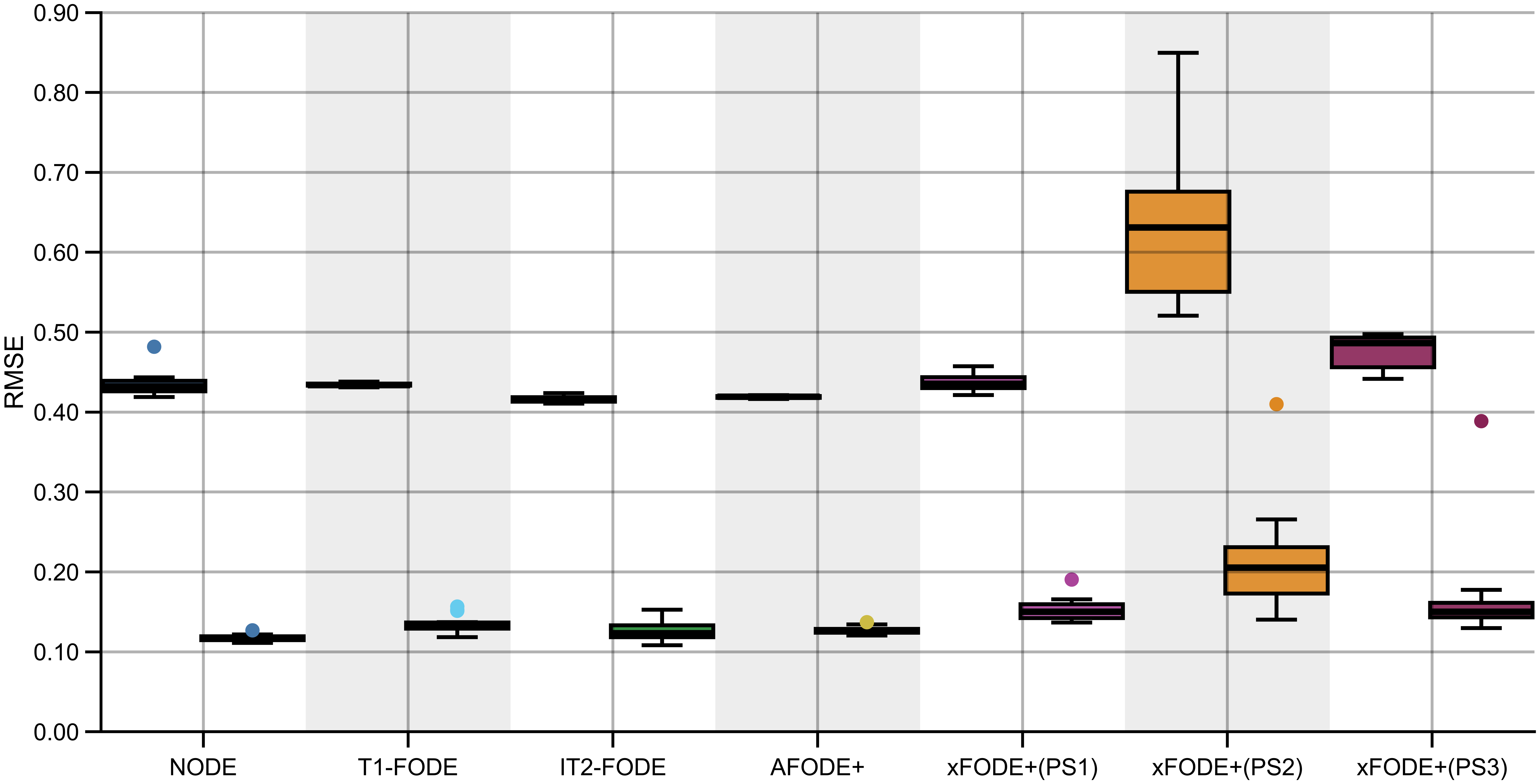}\label{fig:hdrmse}
          }
          \subfigure{   \includegraphics[width=0.46\textwidth,height=0.21\textheight,keepaspectratio]{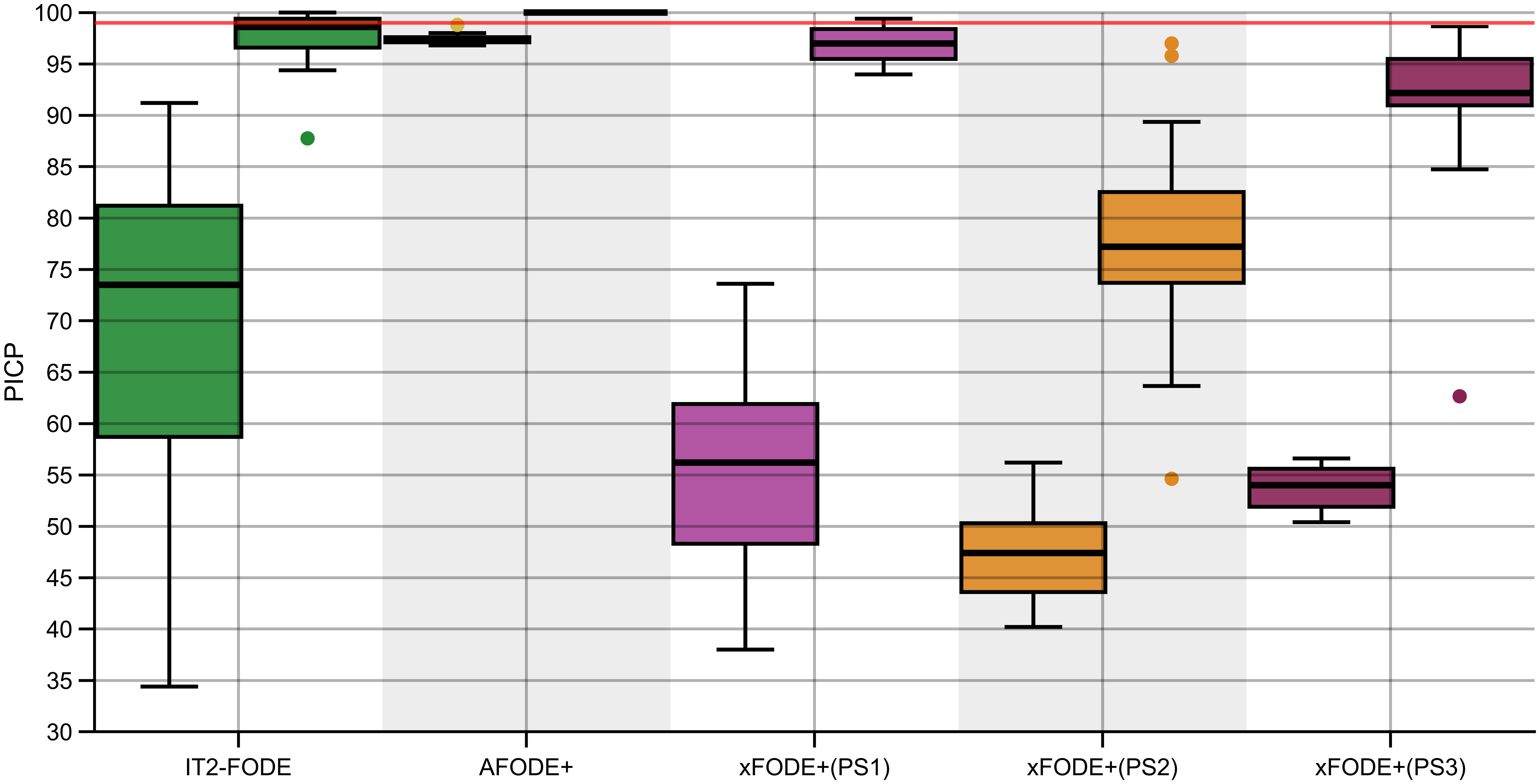}\label{fig:hdpicp}
          }
        \caption{RMSE and PICP boxplots of trained models on Hair Dryer dataset. For each model, the left boxplot corresponds to SR0, and the right boxplot corresponds to SR1. The horizontal red line depicts the desired coverage. }
        \label{fig:hd}
\end{figure}

\begin{figure*}[t]
        \centering
        \subfigure
        {
        \includegraphics[width=0.43\textwidth,height=0.21\textheight,keepaspectratio]{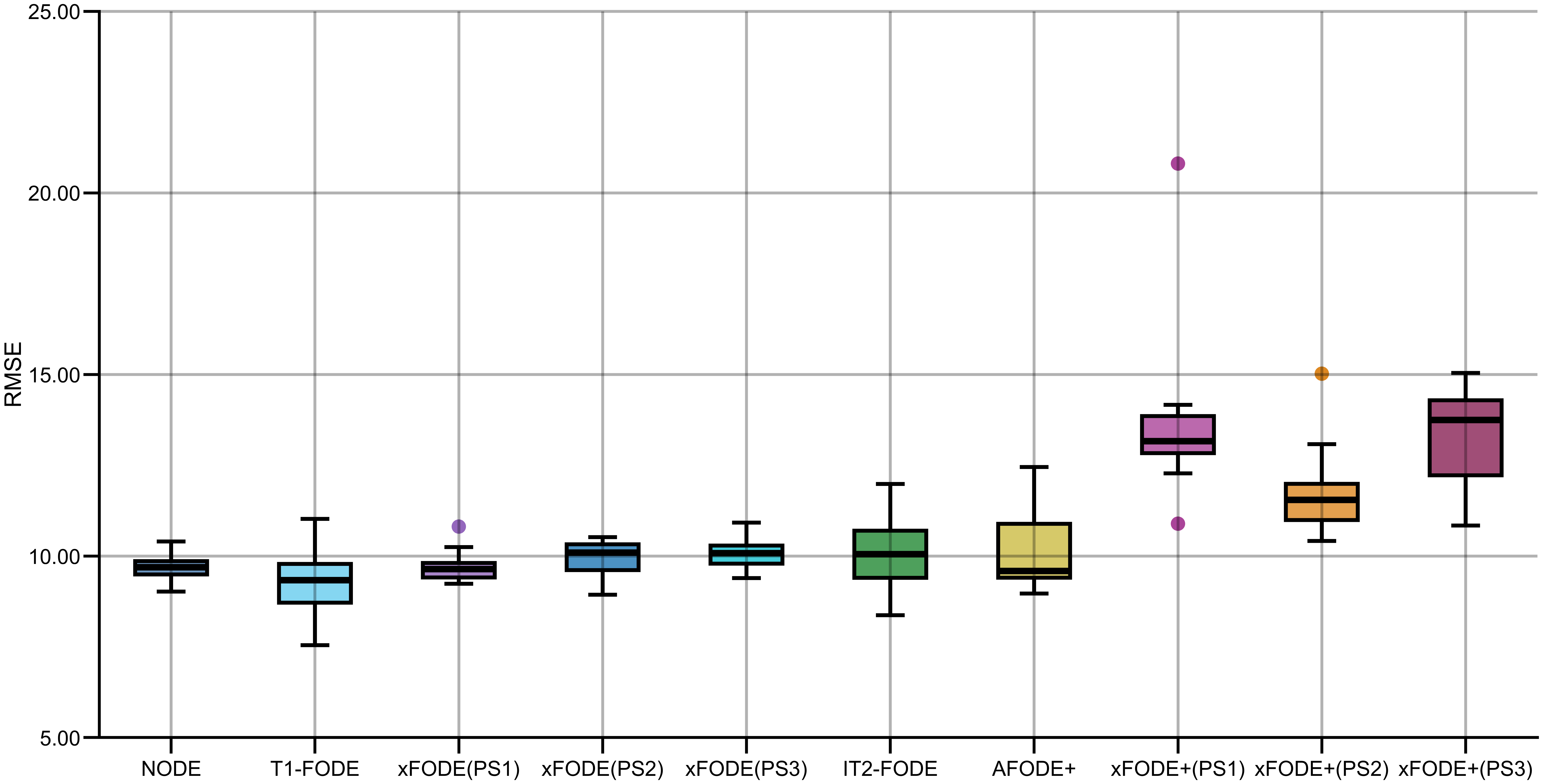}\label{fig:mrdmse}

        }
        \subfigure
        {
        \includegraphics[width=0.43\textwidth,height=0.21\textheight,keepaspectratio]{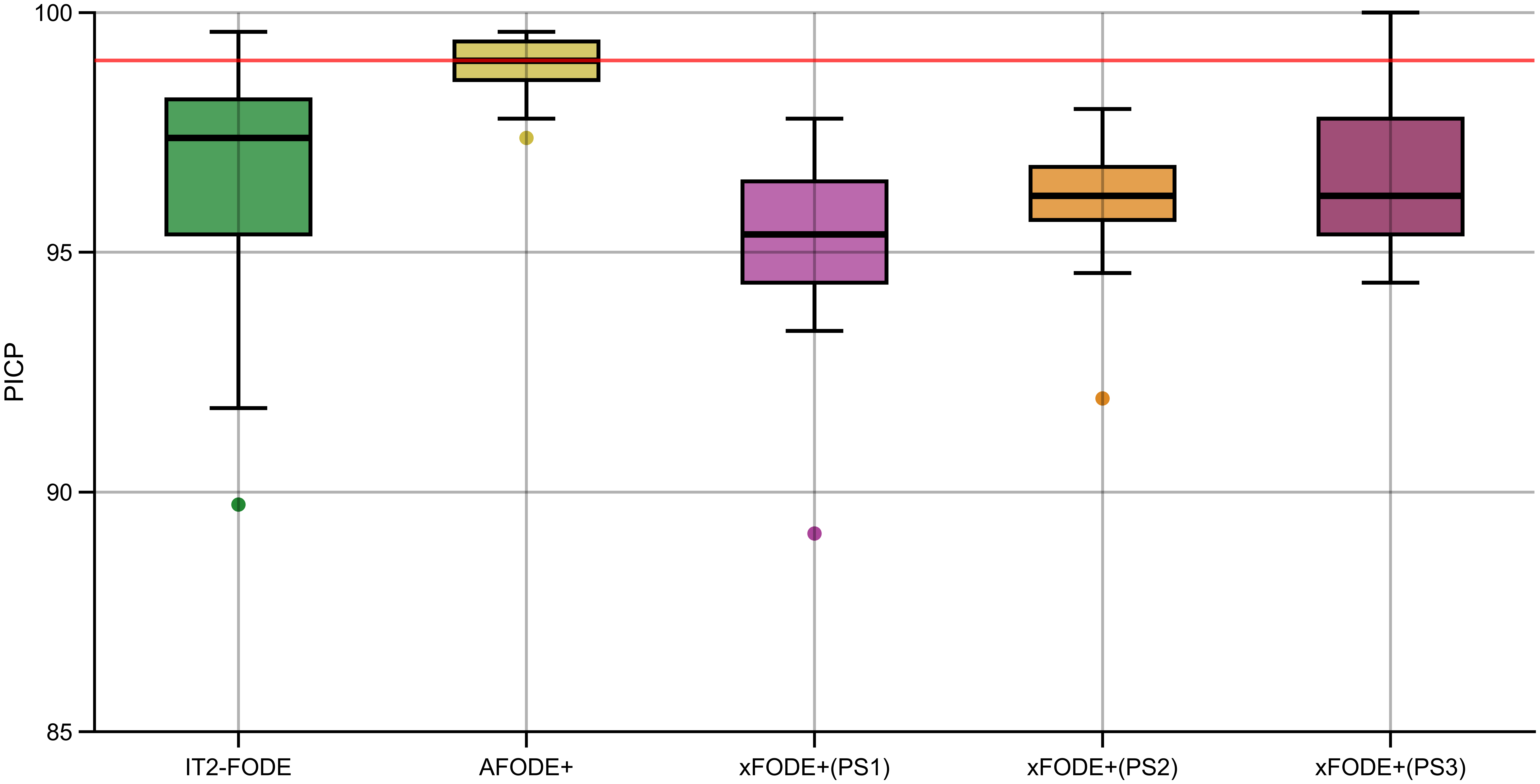}\label{fig:mrdpicp}
        
        }
        \caption{RMSE and PICP boxplots of trained models on the MR Damper dataset. The horizontal red line depicts the desired coverage.}
        \label{fig:mrd}
\end{figure*}

\begin{figure*}[t]
        \centering
        \subfigure
        {
        \includegraphics[width=0.43\textwidth,height=0.21\textheight,keepaspectratio]{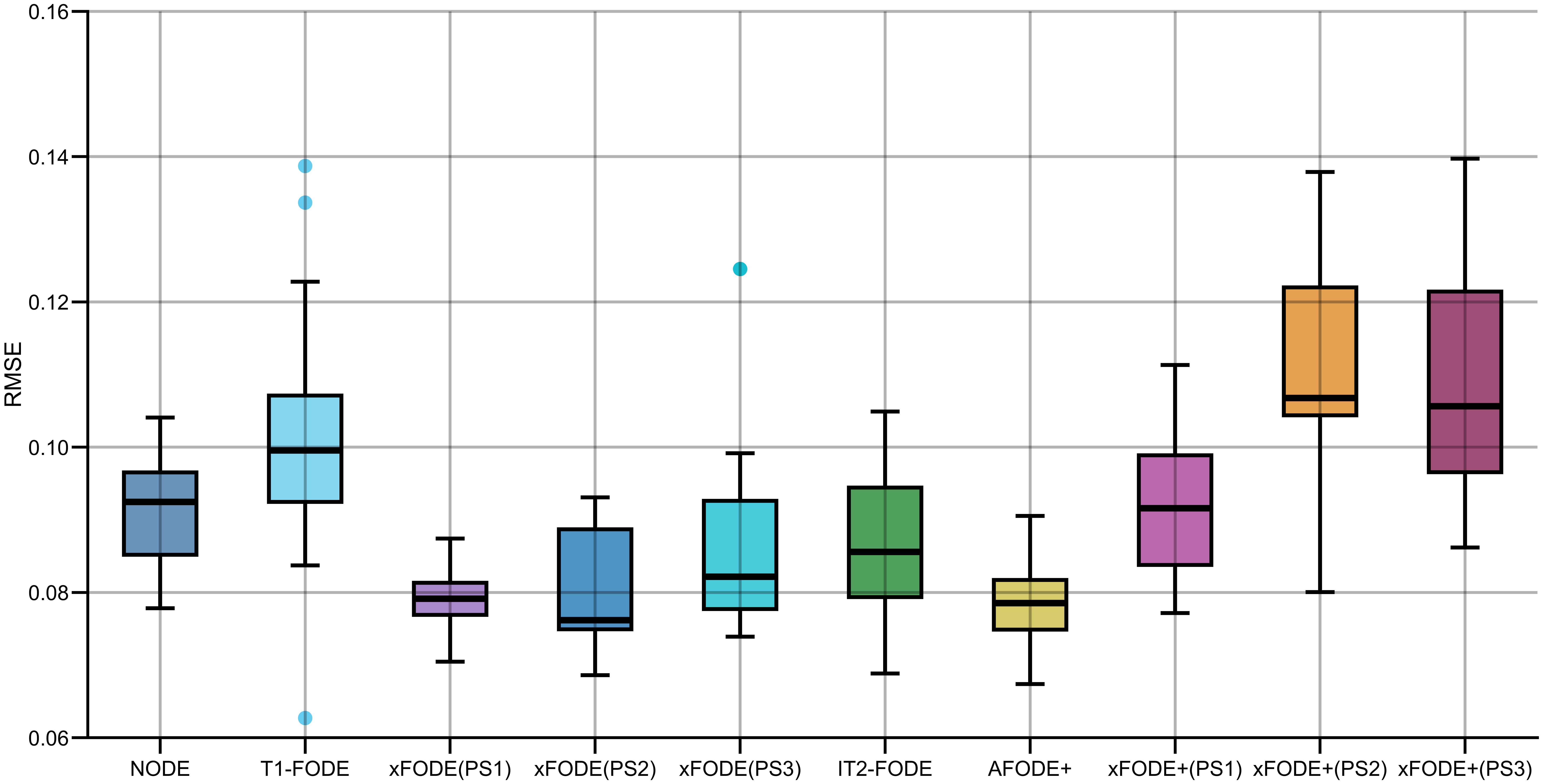}

        }
        \subfigure
        {
        \includegraphics[width=0.43\textwidth,height=0.21\textheight,keepaspectratio]{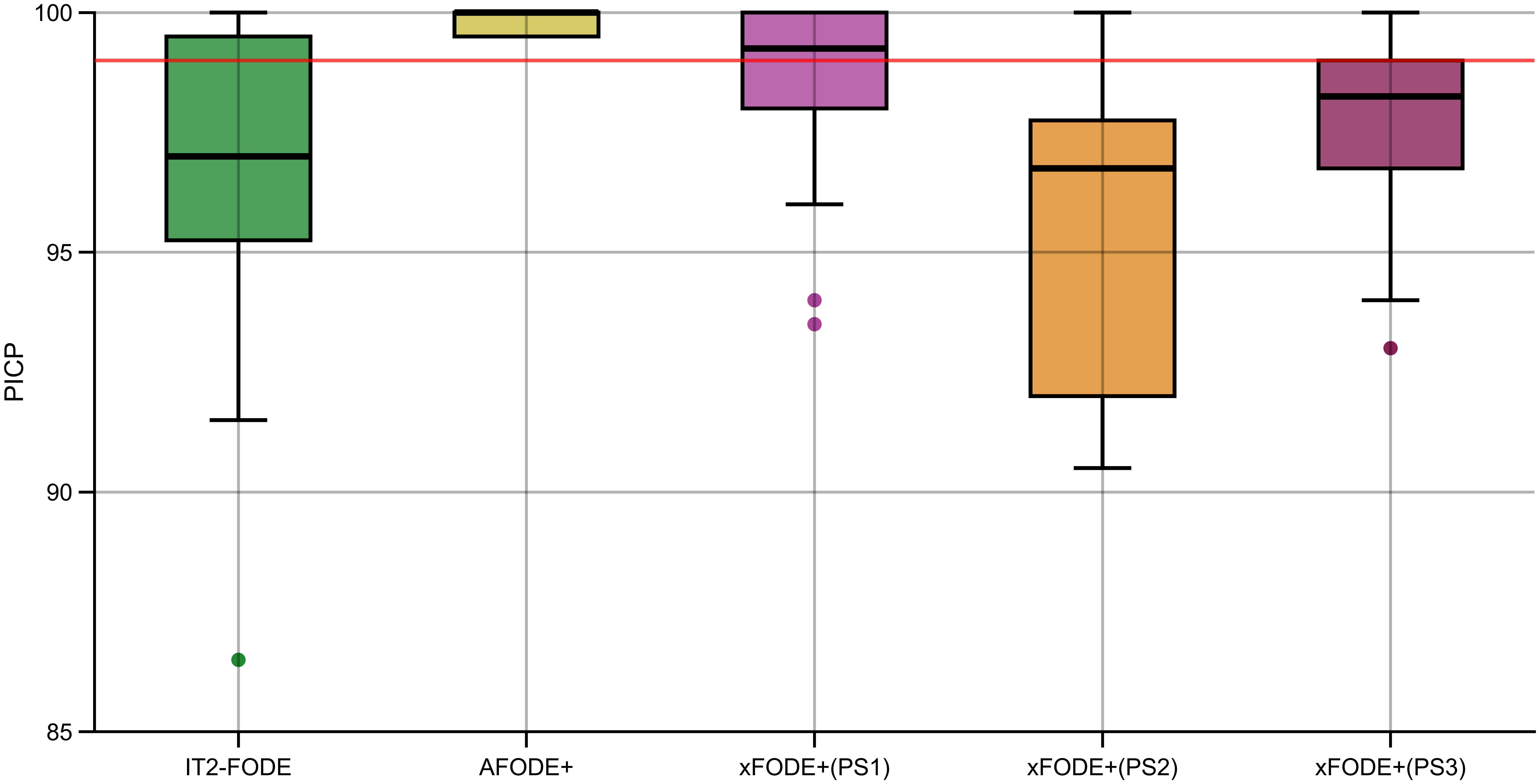}
        
        }

        \subfigure
        {
        \includegraphics[width=0.43\textwidth,height=0.21\textheight,keepaspectratio]{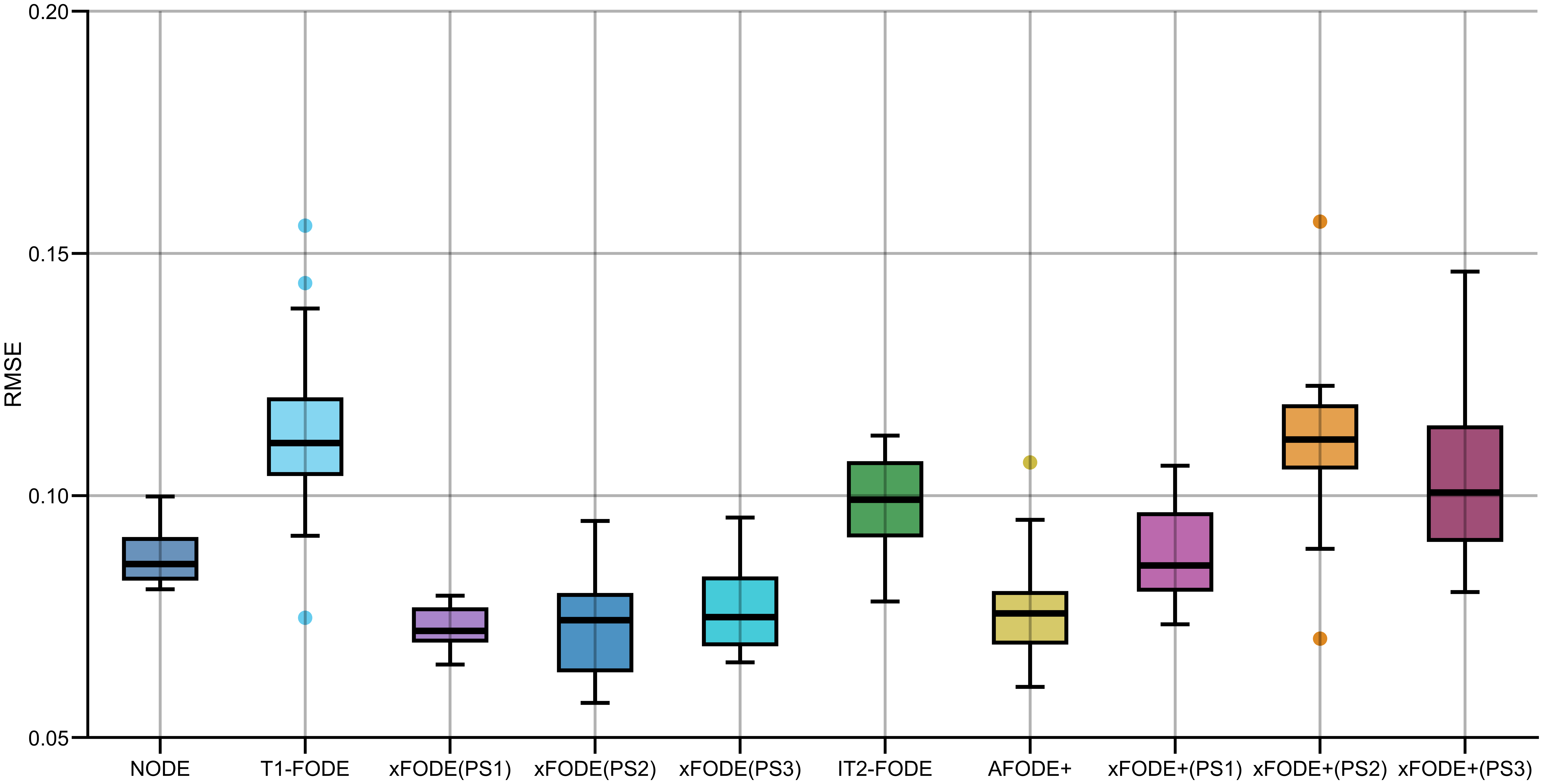}
        }
        \subfigure
        {
        \includegraphics[width=0.43\textwidth,height=0.21\textheight,keepaspectratio]{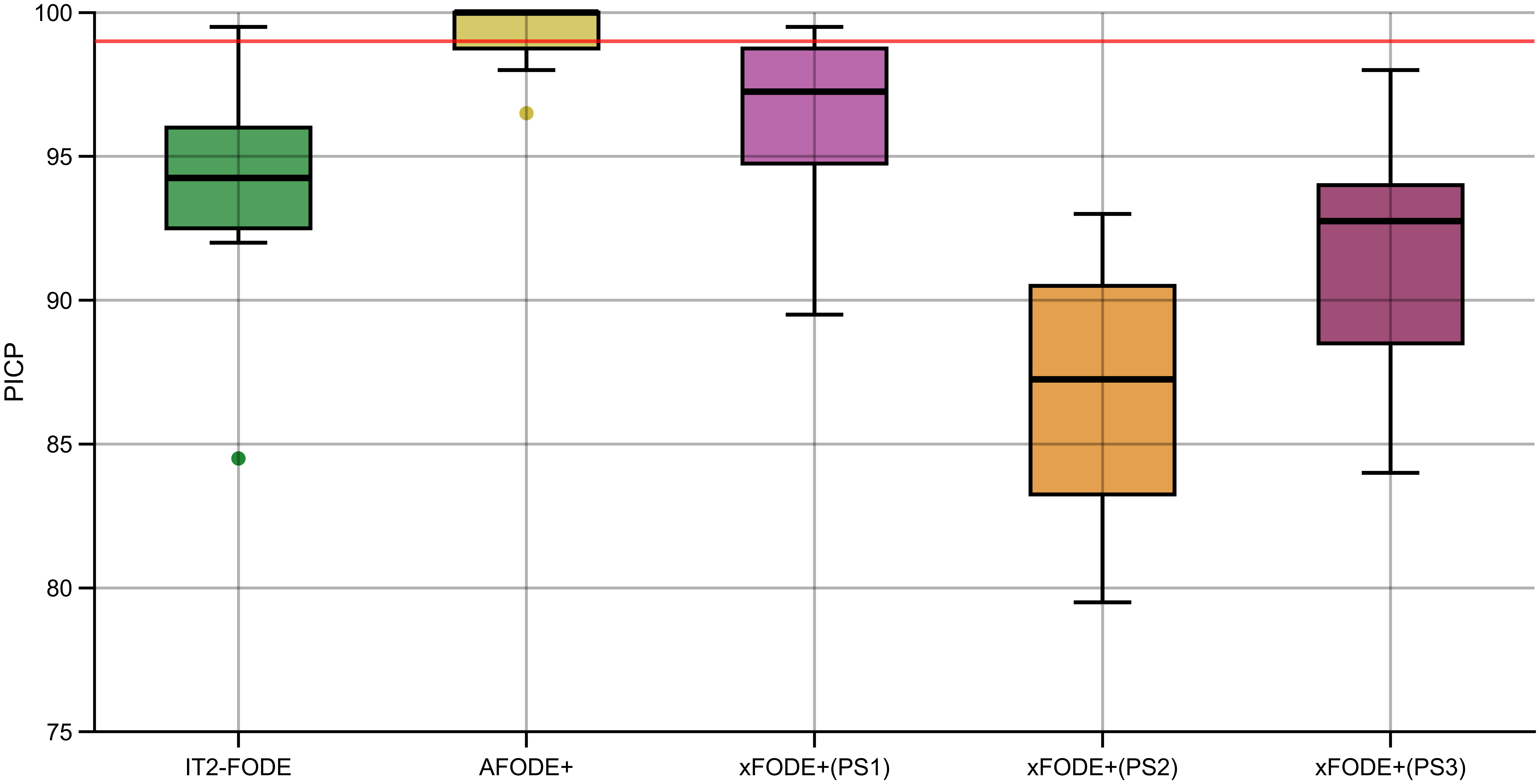}
        
        }
        \caption{RMSE and PICP boxplots of trained models on the Steam Engine dataset; top row corresponds to $y_1$ and bottom row corresponds to $y_2$. The horizontal red line depicts the desired coverage.}
        \label{fig:steam}
\end{figure*}

We observe that models trained with the SR0 consistently perform poorly across all datasets, as illustrated for the Hair Dryer dataset in Fig.~\ref{fig:hd}. Thus, we focus exclusively on SR1 models. Table~\ref{tab:perf_models_merged} reports the corresponding mean RMSE, uncertainty metrics, and \#LP. We conclude that: 
\begin{itemize}
    \item xFODE+ generates PIs comparable to IT2-FODE on MR Damper and Steam Engine across all PSs (Figs. \ref{fig:mrd}--\ref{fig:steam}), with slightly wider PIs, while achieving similar RMSEs. 
    \item AFODE+ yields PIs that achieve target coverage, while obtaining low RMSE. However, its learned MFs may overlap (Fig. \ref{fig:gaussmf}), whereas xFODE+ restricts activation to two neighboring MFs (Figs. \ref{fig:trimf}–\ref{fig:gauss2mf-2}).
    \item NODE and T1-FODE achieve slightly lower RMSE, but they are optimized for accuracy only. xFODE+ instead optimizes a composite loss that also accounts for PI quality with far fewer \#LP than NODE.
    \item xFODE shows superior SysID performance than xFODE+, indicating a trade-off between predictive accuracy and UQ capability introduced in xFODE+.
    \item On the MR Damper dataset, xFODE+ yields similar PIs across all PSs. On the Hair Dryer and Steam Engine datasets, PS1 gives PIs closest to the target coverage compared to PS2 or PS3.
\end{itemize}
Overall, xFODE+ enhances interpretability through structured PSs while maintaining PI quality comparable to IT2-FODE, with similar RMSE and \#LP.

\section{Conclusion and Future Work}

This paper introduced xFODE+, an explainable SysID framework that extends xFODE with UQ by leveraging IT2-FLSs to generate PIs. It represents states in incremental form to preserve physical meaning, and aggregates the TRSs from each IT2-FLS to compute the state update and its PI. Using structured antecedent partitioning, xFODE+ yields transparent IT2-FSs with clean linguistic semantics (e.g., Negative–Zero–Positive) (Fig. \ref{fig:MFs}). While xFODE+ offers such an enhancement, it maintains accuracy and UQ performance comparable to IT2-FODE, with a similar \#LP.

Future work will focus on analyzing the interpretability of the rule consequents and extending the evaluation of xFODE+ to a broader range of datasets.

\begin{table*}[t]
\centering
\caption{Mean Testing performance metrics over 20 experiments}
\label{tab:perf_models_merged}
\renewcommand{\arraystretch}{0.99}
\setlength{\tabcolsep}{4pt}
\footnotesize
\begin{threeparttable}
\begin{tabular}{|l|c|c|c|c|c|c|c|c|c|c|c|}
\hline
\multirow{2}{*}{\textbf{Model}}
& \multicolumn{4}{c|}{\textbf{Hair Dryer $(n_u=1,\,n_y=1)$}}
& \multicolumn{4}{c|}{\textbf{MR Damper $(n_u=1,\,n_y=1)$}}
& \multicolumn{3}{c|}{\textbf{Steam Engine $(n_u=2,\,n_y=2)$}} \\
\cline{2-12}
& \#LP & RMSE & PICP & PINAW
& \#LP & RMSE & PICP & PINAW
& \#LP & RMSE & PICP / PINAW \\ \hline

NODE & 17539 & 0.1173 & -- & --
         & 17539 & 9.6809 & -- & --
         & 17924
         & \begin{tabular}{@{}c@{}}$y_1$: 0.0909\\$y_2$: 0.0873\end{tabular}
         & \begin{tabular}{@{}c@{}}$y_1$: -- / --\\$y_2$: -- / --\end{tabular} \\ \hline

FODE & 115 & 0.1352 & -- & --
         & 115 & 9.2819 & -- & --
         & 200
         & \begin{tabular}{@{}c@{}}$y_1$: 0.1014\\$y_2$: 0.1129\end{tabular}
         & \begin{tabular}{@{}c@{}}$y_1$: -- / --\\$y_2$: -- / --\end{tabular} \\ \hline

xFODE(PS1) & 148 & 0.1271 & -- & --
             & 148 & 9.6841 & -- & --
             & 282
             & \begin{tabular}{@{}c@{}}$y_1$: 0.0790\\$y_2$: 0.0729 \end{tabular}
             & \begin{tabular}{@{}c@{}}$y_1$: -- / --\\$y_2$: -- / --\end{tabular} \\ \hline

xFODE(PS2) & 148 & 0.1249 & -- & --
             & 148 & 9.9384 & -- & --
             & 282
             & \begin{tabular}{@{}c@{}}$y_1$:  0.0802\\$y_2$: 0.0740\end{tabular}
             & \begin{tabular}{@{}c@{}}$y_1$: -- / --\\$y_2$: -- / --\end{tabular} \\ \hline

xFODE(PS3) & 148 & 0.1293 & -- & --
             & 148 & 10.0630 & -- & --
             & 282
             & \begin{tabular}{@{}c@{}}$y_1$: 0.0862\\$y_2$: 0.0768\end{tabular}
             & \begin{tabular}{@{}c@{}}$y_1$: -- / --\\$y_2$: -- / --\end{tabular} \\ \hline

IT2-FODE & 135 & 0.1258 & 97.60 & 0.270
             & 135 & 10.1466* & 96.56 & 0.417
             & 230
             & \begin{tabular}{@{}c@{}}$y_1$: 0.0864\\$y_2$: 0.0979\end{tabular}
             & \begin{tabular}{@{}c@{}}$y_1$: 96.72 / 0.247\\$y_2$: 94.10 / 0.683\end{tabular} \\ \hline

AFODE+ & 180 & 0.1268 & 100.00 & 0.431
           & 180 & 10.2086 & 98.81 & 0.411
           & 330
           & \begin{tabular}{@{}c@{}}$y_1$: 0.0785\\$y_2$: 0.0770\end{tabular}
           & \begin{tabular}{@{}c@{}}$y_1$: 99.85 / 0.242\\$y_2$: 99.35 / 0.732\end{tabular} \\ \hline

% PS1 is old PS3
xFODE+(PS1) & 168 & 0.1524 & 96.84 & 0.310
              & 168 & 13.5024 & 95.24 & 0.418
              & 312
              & \begin{tabular}{@{}c@{}}$y_1$: 0.0917\\$y_2$: 0.0879\end{tabular}
              & \begin{tabular}{@{}c@{}}$y_1$: 98.55 / 0.276\\$y_2$: 96.27 / 0.679\end{tabular} \\ \hline

% PS2 is old PS1
xFODE+(PS2) & 168 & 0.2092 & 77.92 & 0.267
              & 168 & 11.6543 & 96.07 & 0.452
              & 312
              & \begin{tabular}{@{}c@{}}$y_1$: 0.1100\\$y_2$: 0.1105\end{tabular}
              & \begin{tabular}{@{}c@{}}$y_1$: 95.40 / 0.270\\$y_2$: 86.97 / 0.591\end{tabular} \\ \hline

% PS3 is old PS2
xFODE+(PS3) & 168 & 0.1625 & 91.41 & 0.317
              & 168 & 13.2810 & 96.59 & 0.489
              & 312
              & \begin{tabular}{@{}c@{}}$y_1$: 0.1096\\$y_2$: 0.1034\end{tabular}
              & \begin{tabular}{@{}c@{}}$y_1$: 97.55 / 0.308\\$y_2$: 91.92 / 0.689\end{tabular} \\ \hline

\end{tabular}
\begin{tablenotes}
\footnotesize
\scriptsize
\item[*] IT2-FODE yielded NaN values for 2 seeds. These experiments are excluded from statistical analysis.
\end{tablenotes}
\end{threeparttable}
\end{table*}

% \begin{figure}[t] 
% \includegraphics[width=\linewidth]{Figs/HairDryer_PINAW_col1_BOTH_RAW.png}
% \caption{PINAW boxplots of trained models on Hair Dryer dataset; SR0 with the white, SR1 with the gray background.}
%     \label{fig:hdpinaw}
% \end{figure}

% \begin{figure}[t] 
% \includegraphics[width=\linewidth]{Figs/MRDamper_PINAW_col1_SR1_RAW.png}
% \caption{PINAW boxplots of trained models with SR1 on the MR Damper dataset.}
%     \label{fig:mrdpinaw}
% \end{figure}

% \begin{figure}[t] 
% \includegraphics[width=\linewidth]{Figs/SteamEngine_PINAW_col2_SR1_RAW.png}
% \caption{PINAW boxplots of $y_2$ of trained models with SR1 on the Steam Engine dataset.}
%     \label{fig:stepinaw2}
% \end{figure}

\section*{Acknowledgment}
The authors acknowledge using ChatGPT to refine the grammar and enhance the expression of English.

\bibliographystyle{IEEEtran}
\bibliography{IEEEabvr,cites}

 % \addtolength{\textheight}{-20cm}   % This command serves to balance the column lengths

%%%%%%%%%%%%%%%%%%%%%%%%%%%%%%%%%%%%%%%%%%%%%%%%%%%%%%%%%%%%%%%%%%%%%%%%%%%%%%%%

%%%%%%%%%%%%%%%%%%%%%%%%%%%%%%%%%%%%%%%%%%%%%%%%%%%%%%%%%%%%%%%%%%%%%%%%%%%%%%%%

%%%%%%%%%%%%%%%%%%%%%%%%%%%%%%%%%%%%%%%%%%%%%%%%%%%%%%%%%%%%%%%%%%%%%%%%%%%%%%%%

\end{document}